\documentclass[11pt,a4paper]{article}
\usepackage{authblk}
\usepackage[hyperref]{eacl2021}
\usepackage{times}
\usepackage{latexsym}
\usepackage[normalem]{ulem}
\usepackage{svg}
\usepackage{subcaption}
\usepackage{textcomp}
\usepackage{paralist}
\usepackage{multirow}

\usepackage{graphicx}
\usepackage{microtype}
\usepackage{makecell}
\usepackage{graphicx}
\usepackage{multirow}
\usepackage{hyperref}
\usepackage{xcolor,colortbl}

\aclfinalcopy 

\title{Leveraging Commonsense Knowledge on Classifying False News and Determining Checkworthiness of Claims}

\author[1]{Ipek Baris Schlicht\thanks{Corresponding Author: ibarsch@doctor.upv.es. Work done during the time at the University of Koblenz-Landau, Germany.}}
\author[2]{Erhan Sezerer}
\author[2]{Selma Tekir}
\author[3]{Oul Han}
\author[4]{Zeyd Boukhers}
\affil[1]{ibarsch@doctor.upv.es, Universitat Politècnica de València / Valencia, Spain}
\affil[2]{\{erhansezerer,selmatekir\}@iyte.edu.tr\\Izmir Institute of Technology / Izmir, Turkey}
\affil[3]{hanoul@gmail.com, 2bytes Corp. / Anyang, South Korea}
\affil[4]{boukhers@uni-koblenz.de, University of Koblenz-Landau / Koblenz, Germany}

\definecolor{Gray}{gray}{0.95}
\newcolumntype{g}{>{\columncolor{Gray}}c}
\date{}

\newif\ifproofread

\newcommand{\changemarker}[1]{%
\ifproofread
\textcolor{red}{#1}%
\else
#1%
\fi
}

\begin{document}
\maketitle
\begin{abstract}
Widespread and rapid dissemination of false news has made fact-checking an indispensable requirement. Given its time-consuming and labor-intensive nature, the task calls for an automated support to meet the demand. In this paper, we propose to leverage commonsense knowledge for the tasks of false news classification and check-worthy claim detection. Arguing that commonsense knowledge is a factor in human believability, we fine-tune the BERT language model with a commonsense question answering task and the aforementioned tasks in a multi-task learning environment. For predicting fine-grained false news types, we compare the proposed fine-tuned model's performance with the false news classification models on a public dataset as well as a newly collected dataset. We compare the model's performance with the single-task BERT model and a state-of-the-art check-worthy claim detection tool to evaluate the check-worthy claim detection. Our experimental analysis demonstrates that commonsense knowledge can improve performance in both tasks.
\end{abstract}
\vspace{-0.4cm}
\section{Introduction}
\proofreadfalse
The increase of social media usage in recent years has changed the way of consuming news. Although social media is useful for following the updates of breaking events such as the COVID-19 pandemic, it also misleads through false news spreading rapidly and globally on platforms \citep{KarlovaF13,VosoughiSpread2018}, which results in negative emotions, confusion, and anxiety in society \citep{BudakAA11} and even in the manipulation of major outcomes, such as political elections \citep{VosoughiSpread2018, LazerScience2018}. 

To combat false news, the number of fact-checking initiatives around the world has increased. However, manual fact-checking can satisfy only a small share of the demand as fact-checking is a labor intensive and time consuming task. It requires fact checkers to contact people/organizations that are mentioned in the claim, consult experts that provide background knowledge if needed, seek source validation, etc. Typical fact-checking for one news item takes about one day in order to research the facts and report the result \citep{GravesUnderstanding2018}. This yields a time lag between the spread of false news and the delivery of the fact-checked article  \citep{HassanQuest2015}.

In recent years, we have seen a growing interest in developing computational approaches for combating false news. Some studies focus on automatising the steps of manual fact-checking \citep{CazalensLLMT18,ThorneV18,GravesUnderstanding2018}. As a seminal study, ClaimBuster \citep{hassan2015detecting,HassanZACJGHJKN17} is the first end-to-end automatic fact-checking framework and is widely used by professional fact checkers and journalists \citep{adair2019human}. ClaimBuster identifies check-worthy factual claims in texts from various data sources (e.g social media, websites) using a classification and score model that is trained on human labeled political debates, searches identified claims against fact-checking websites, collects supporting/debunking evidences and in a final step creates a report which combines the search results, aforementioned evidences and claim checkworthiness score. Although ClaimBuster is able to spot simple declarative claims, it misses those implicitly stated in sentences \citep{GravesUnderstanding2018}.

Other studies focus on false news detection based on the style of news content or social media context \citep{ShuSWTL17,zhou2018survey}. Most of these studies treat false news detection as a binary classification problem that labels a news article as fake or true. As an example, \citet{SinghaniaFR17} propose a three level hierarchical attention network model (3HAN) that exploits the article structure. The authors evaluated on a dataset that is constructed of a small list of fake and legit news sources and outperformed models such as single level hierarchical attention network (HAN) \citep{YangYDHSH16}.  

However, on the web, multiple types of false news can be found such as satire or propaganda \citep{ZannettouSBK19,RashkinCJVC17}. While satire conveys irony or parody and contains unrealistic situations, propaganda mimics the style of real news and can mislead readers with malintent. Furthermore, political polarization propels biases in news \citep{LazerScience2018}.  False news detection models must therefore be more fine-grained than the aforementioned binary models in order to address diverse types of false news. Therefore, the literature is in quest of new techniques and methodologies \citep{BozarthB20} as well as datasets \citep{torabi2019big}.

Addressing the diversity of false news, we identify one common trait that enables them all. We point out that the impression of veracity is created by a seeming plausibility of news stories, since humans believe depending on how much a story fits their prior knowledge \citep{connell2004plausibly,connell2003pam}. This background knowledge is termed as \textit{commonsense knowledge}. \citet{GoldwasserZ16} already demonstrated that commonsense knowledge can improve satire detection better than traditional text classification models. With this motivation, we integrate commonsense knowledge into the tasks of false news classification and check-worthy claim detection, applied on diverse news. 
\\
\begin{table*}[ht]
    \centering
    \small
    \begin{tabular}{|p{15cm}|}
    \hline
         \textbf{Task A: Commonsense Question Answering (CSQA)}  \\
         \hline
          \textbf{Input 1: Question} People are very much like the animals, but one thing has secured or dominance over the planet.  We're better at doing what? \\
          \textbf{Input 2: Multiple Choices} a) Eat eggs b) Make tools c) eat dosa d) talk to each other e) smoke pot \\
          \textbf{Output: Answer} b \\
     \hline 
     \textbf{Task B: False News Classification} \\
     \hline
     \textbf{Input 1: Title} TSA Agents Can Now Grope Travelers Without Fear Of Pesky Lawsuits \\
     \textbf{Input 2: Body} Transportation Security Administration (TSA) screeners have gained the upper glove when it comes  to being sue ... \\
     \textbf{Output: News Type} Conspiracy \\
     \hline
     \textbf{Task C: Check-worthy Claim Detection} \\
     \hline
     \textbf{Input: Statement} And I saw the cocaine scene treated with humor, as though this was a humorous little incident. \\
     \textbf{Output: Checkworthiness} Unimportant Factual Statement (UFS)\\
     \hline
    \end{tabular}
    \caption{Examples from each task that shows inputs and outputs.}
    \vspace{-0.6cm}
    \label{tab:task_definitions}
\end{table*}
In our study, commonsense knowledge is captured by a commonsense question answering (CSQA) task \citep{TalmorHLB19} where the inputs are questions and multiple choices as answer candidates, and the output is an answer (Task A in Table~\ref{tab:task_definitions}). For classifying different types of news on the web, we leverage the false news taxonomy proposed by \citet{ZannettouSBK19}. In this task, the inputs are body and title of a news article and the output is the news type, namely satire, conspiracy, propaganda, bias-right, bias-left or neutral (Task B in Table~\ref{tab:task_definitions}). In check-worthy claim detection, the input is a sentence and the output is a label indicating its checkworthiness, namely, check-worthy factual statement (CFS), not factual statement (NFS) and unimportant factual statement (UFS) (Task C in Table~\ref{tab:task_definitions}). To transfer commonsense knowledge to task B and C using multitask training \citep{LiuHCG19}, we prepend them with fine-tuning a pre-trained BERT model over a CSQA task.

The main contributions of this study are as follows:
\begin{compactitem}
    \item To the best of our knowledge, it is the first attempt to leverage commonsense knowledge \changemarker{to classify fake news and detect check-worthy claims.} 
    \item We collected a new community interest news dataset (CIND) from the social media platform Reddit. Unlike the publicly available false news collection NELA 2019 \citep{NELA2019}, CIND is a collection (1) of news that news consumers found interesting or plausible, (2) that features a diverse number of sources which makes the experiment of predicting news from an unseen source more reliable (3) that covers news events which occurred from 2016 to 2019, allowing for forecasting. We label both datasets with the false news taxonomy \citep{ZannettouSBK19}.
    \item We conducted an extensive set of experiments to validate our hypothesis. The results show that commonsense knowledge \changemarker{could} improve the (1) predictions of 4 out of 6 classes on CIND and the predictions of bias-right articles from the NELA dataset for the experiment of predicting news articles from unseen sources, (2) the predictions of 3 out of 6 classes on CIND for the experiment of forecasting, (3) the prediction of all classes in check-worthy claim detection task.
\end{compactitem}


The rest of the paper is organized as follows. Section~\ref{related_work} summarizes related work. Section~\ref{models} presents the proposed models. Section \ref{datasets} introduces the new dataset collected for this study along with other datasets used for comparison. Finally, section~\ref{results} discusses the experimental results.
\vspace{-0.2cm}
\section{Related Work\label{related_work}} 
In this section, we present the studies related to our research. Section~\ref{check-worthyclaimdetection} and section~\ref{falsenewsdetection} outline the studies in check-worthy claim detection and false news classification, section~\ref{commonsense-studies} presents the studies encoding commonsense knowledge for text classification tasks and section~\ref{multi-tasking-related-works} presents the studies leveraging multi-task learning for fact-checking and false news classification.  

\vspace{-0.2cm}
\subsection{Check-worthy Claim Detection\label{check-worthyclaimdetection}}
Check-worthy claim detection is the first step of the fact-checking pipeline \citep{CazalensLLMT18,GravesUnderstanding2018,ThorneV18}. The component of ClaimBuster \citep{hassan2015detecting,HassanZACJGHJKN17} that detects check-worthy claims is trained with a SVM classifier using tf-idf bag of words, named entity types, POS tags, sentiment and sentence length as a feature set. \citet{GenchevaNMBK17} proposed a fully connected neural network model trained on claims and their related political debate content. Additionally, CLEF Check That! Lab (CTL) has organized shared tasks to tackle this problem in political debates \citep{AtanasovaMBESZK18,AtanasovaNKMM19}  and in social media \citep{barron2020overview}. 
    
\vspace{-0.2cm}
\subsection{Style-based False News Classification\label{falsenewsdetection}}
Style-based approaches for false news classification attempt to capture the writing style or deceptive clues in news articles \citep{ShuSWTL17,zhou2018survey,SteinPKBR18}. The methods range from hand-crafted feature-based methods \citep{VolkovaSJH17,Perez-RosasKLM18} to sophisticated deep neural networks \citep{SinghaniaFR17, RiedelASR17, KarimiT19}.\\
\citet{RiedelASR17} focus on the first part of the fake news detection problem: stance detection. Their model RDEL extracts the most frequent unigrams and bigrams, constructs tf-idf vectors for article headlines and bodies, and also computes the cosine similarity of headline and body. Finally, all of these features are fed into a Multilayer Perceptron for classification.\\
The model 3HAN \citep{SinghaniaFR17} encodes the body of news articles similar to HAN \citep{YangYDHSH16} where the words in each sentence are encoded with BiGRU \citep{ChoMGBBSB14} and then an attention mechanism \citep{BahdanauCB14} identifies informative words in a sentence. Finally sentences are passed through an attention layer to find informative sentences in the document before classifying them with a dense layer. In addition to HAN, it also concatenates the encoded headline with the processed body of the article and runs attention mechanism on the concatenated features before feeding it to the dense layer.\\
Most of these aforementioned studies tackle the problem as binary classification. Few scholars have investigated different types of false news. \citet{RubinCC15} differentiated false news types as hoaxes, satire and deliberately misleading fabrications. \citet{RashkinCJVC17} studied linguistic features for analysis of hoax, propaganda, satire and trusted sources. \citet{GhanemRP20} analyzed the impact of emotions in clickbaits in addition to the sources that \citet{RashkinCJVC17} studied. However, the dataset used in both studies \citep{rashkin-etal-2017-multilingual, GhanemRP20} covers \changemarker{a few sources} for each category. In our study, we increase the number of sources for each category and include articles from biased sources.  
\vspace{-0.2cm}
\subsection{Incorporating Commonsense Knowledge\label{commonsense-studies}}
Incorporating commonsense knowledge into text representations can improve many tasks in NLP and NLU, such as machine comprehension (i.e \citet{Wang2018YuanfudaoAS}), question answering (i.e \citet{ZhongTDZWY19}) and opinion mining (i.e \citet{du2020commonsense,xu2019opinion,ZhongWM19,MaPC18}. Closest to this paper is the study by \citet{GoldwasserZ16} who leverage commonsense knowledge for satire detection. Their approach constructs a narrative representation of an article by extracting main actors, events and statements, then makes inferences to quantify the likelihood of those entities appeared in a real/satire context.\\
 To leverage commonsense knowledge, one approach is to use knowledge aware distributional word embeddings such as Numberbatch \citep{speer2017conceptnet} which is built on the commonsense knowledge base ConceptNet. Alternatively, commonsense knowledge can be transferred by using multi-task learning \citep{BosselutRSMCC19}. In our study, we evaluate both approaches.

\vspace{-0.2cm}
\subsection{Multi-task Learning\label{multi-tasking-related-works}}
Multi-task learning is motivated by human learning. While learning new tasks, we apply the knowledge that is gained from related tasks. In contrast to single-task learning, multi-task learning can learn a more general representation by leveraging the knowledge of auxiliary tasks when the original task has noisy/small amount of samples \citep{Ruder17a}.
\par
There are several attempts to apply multi-task learning to the tasks that aid fact-checking or detect false news. \citet{KochkinaLZ18} applied a multi-task learning model that encodes inputs with a shared LSTM and then jointly learns the tasks in a rumour verification pipeline (stance detection, veracity and identifying rumours). \citet{BalyKSGN19} proposed a multi-task ordinal regression framework for jointly predicting the factuality and political ideology of news media. \citet{AtanasovaSLA20} presented a multi-task model for veracity explanation generation and veracity prediction based on DistilBERT\citep{abs-1910-01108}. 
\section{Models\label{models}}
\begin{figure}[t]
    \centering
    \includegraphics[width=\columnwidth]{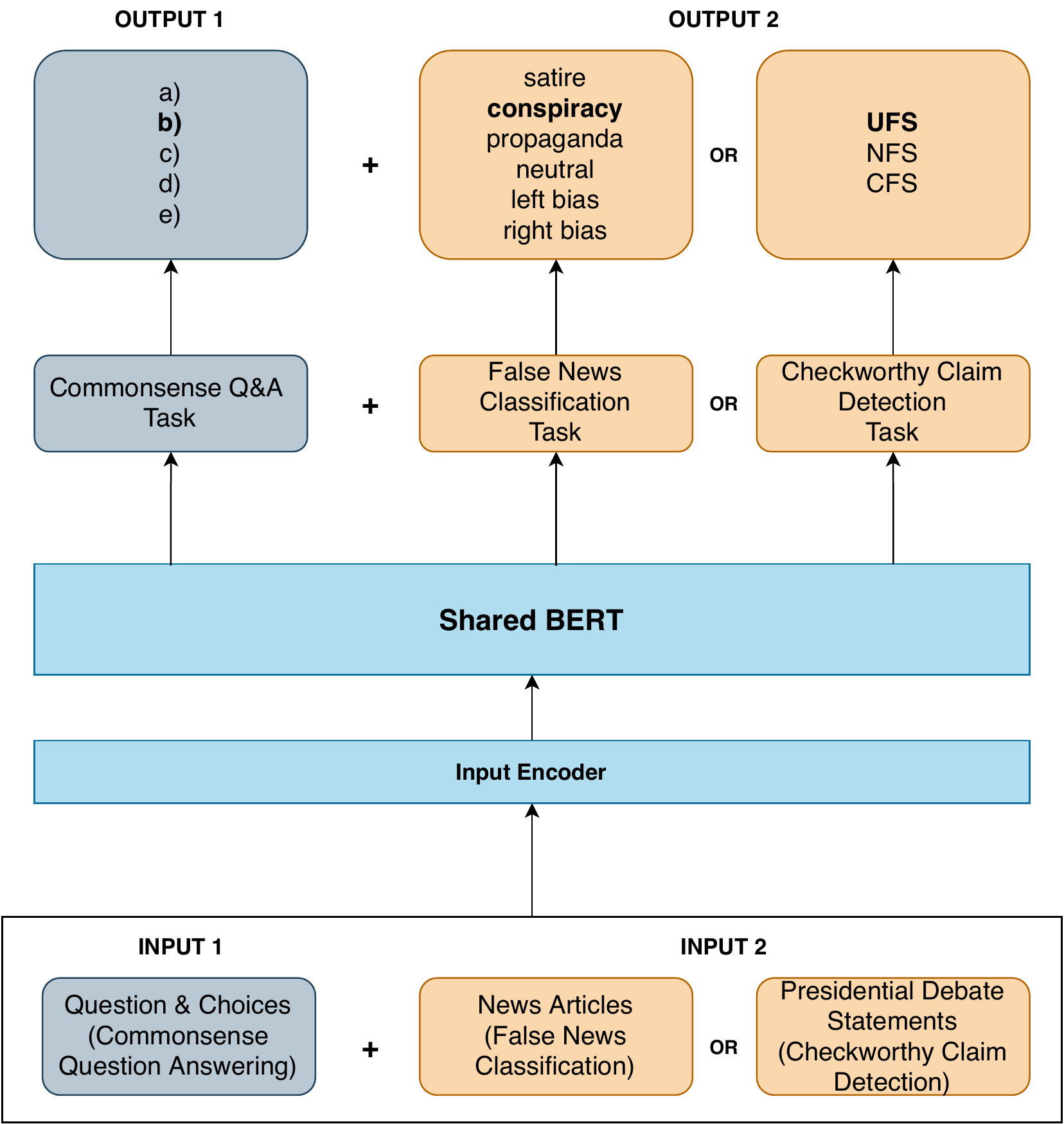}
    \caption{Overview of MTBERT model for false news classification and check-worthy claim detection.}
        \vspace{-0.6cm}
    \label{fig:multitask_bert}
\end{figure}
To test whether the exploitation of commonsense knowledge improves false news classification and check-worthy claim detection, the MTBERT model of \citet{LiuHCG19} was used. Figure~\ref{fig:multitask_bert} illustrates the resulting model architecture. The model consists of two shared lower levels and two task specific layers. Lower layers incorporate the original BERT architecture where the first layer maps the inputs to tokens required by BERT and the second layer resides the transformer encoders of BERT. The upper two layers represent specific task configurations along with their loss functions. The model is trained in two steps: First, the shared BERT model is trained on the pre-training tasks of masked word prediction and next sentence prediction. In the second phase, all samples belonging to all tasks are shuffled. Then, each sample is used to train/fine-tune the shared parameters of BERT with respect to the loss functions of the specific tasks. This training scheme enables the model to learn a task by transferring the information gained from other tasks.\\
In our work, instead of training the BERT model from scratch, we use the HuggingFace library \cite{HuggingFace} to obtain the pre-trained BERT model that would result from the first training phase. In the second phase, we fine-tune the pre-trained model using CSQA as the first task and false news classification or check-worthy claim detection as the second task, depending on the objectives. This allows the model to perform false news classification or check-worthy claim detection tasks, using the information gained from the CSQA task.\\
To test the performance of the fine-tuned MTBERT in false news classification, we compare our results against the state-of-the-art false news classification models RDEL and 3HAN. For both models, we encode the inputs with Glove embeddings \cite{PenningtonSM14}, as originally stated in their papers. These embeddings, however, do not incorporate commonsense knowledge. In order to make them comparable, we also feed Numberbatch embeddings that incorporate ConceptNet commonsense knowledge \citep{speer2017conceptnet} into them. We report the baseline performances of HAN and SVM at this task, additionally. Furthermore, we compare the performance against the original BERT to confirm that the performance gains depend not only on the classifier but also on the use of commonsense knowledge. For SVM, we used the 25,000 most frequent unigrams and bigrams as features. In all the experiments involving BERT and MTBERT where the body of the article is merged with the headline, we used the token \texttt{[SEP]} that is acknowledged by the architecture to separate different contexts.\\
For the task of check-worthy claim detection, we compare MTBERT with the results of the ClaimBuster API. The API gives a probability score for the checkworthiness of a claim where we map scores below 0.33 to NFS, above 0.66 to CFS and the rest to UFS.  We also carry out this experiment on (1) a SVM classifier trained on ClaimBuster features (e.g POS tags, entity types, tf-idf, etc.) with 5-fold cross validation and (2) the original BERT as baselines.

\section{Datasets\label{datasets}}

\subsection{Datasets for False News Classification}
\textbf{NELA 2019} \citep{NELA2019} is a publicly available dataset covering mainstream and alternative news sources. The authors scraped news articles from RSS feeds of all its news sources in 2019. NELA 2019 has the following limitations (1) the number of satire and neutral sources is low (2) only events in 2019 are covered (3) the news publishers select which articles to share in their RSS feeds.
\begin{table}
    \centering
    \small
    \begin{tabular}{l}
         \hline
         \textbf{Subreddits} \\
         \hline
         \texttt{TheDonald}\\
         \texttt{new\_right}\\
         \texttt{news}\\
         \texttt{conspiracy} \\
         \texttt{fakenews} \\
         \texttt{nottheonion} \\
         \texttt{neutralnews}\\
         \texttt{pol} \\
         \texttt{politics} \\
         \texttt{satire} \\
         \texttt{savedyouaclick} \\
         \texttt{worldnews} \\
         \hline
    \end{tabular}
    \caption{List of subreddits that are used for this study.}
    \vspace{-0.5cm}
    \label{reddit_lists}
\end{table}

We propose the community interest news dataset \textbf{CIND} to overcome the aforementioned limitations of NELA 2019. Instead of relying on the news sources' selection of articles, it respects the interest of news consumers. As the collection source, we choose Reddit because (1) it is popular in various communities (2) openly accessible and (3) contains articles from a variety of news sources. Reddit users can share news and discuss them in online communities so-called subreddits. Each subreddit has its own discussion theme and moderation system. For instance, users share satire-like news articles from mainstream news sources in \texttt{r/nottheonion} and discuss conspiracy theories in \texttt{r/conspiracy}. Articles from unreliable sources are removed by the moderators of \texttt{r/nottheonion} while only clickbait articles are allowed to be shared in \texttt{r/savedyouaclick}.
\par
We selected subreddits that have been analyzed in \citet{ZannettouCCKLSS17} or \citet{HorneDKA18}, or was used as the source in a dataset before \citep{NakamuraLW20} which were active within 2016-2019. Additionally, we added \texttt{r/fakenews} where false news stories are highlighted and fact-checks are shared, with respect to the specified time frame. Table~\ref{reddit_lists} lists the subreddits that are used for this study. 
\\
We crawled posts with PushshiftAPI \citep{BaumgartnerZKSB20} by ignoring those that are removed by moderators and users. We filtered out the posts whose metadata contain flair link text\footnote{Examples of some link texts are \texttt{Not oniony - Removed}, \texttt{misleading title}, \texttt{Not a news article - Removed}} which is used for posts that violate subreddit rules. We extracted the articles by using Newspaper3k\footnote{https://newspaper.readthedocs.io/en/latest/}. We filtered out the articles which are not in English, non news sources such as Facebook, Youtube, etc. and sources that were not accessible due to technical issues such as \texttt{huffpost.com}.

\begin{table*}
    \small
    \centering
    \begin{tabular}{lll}
    \hline
    \textbf{Type}&\textbf{Title}&\textbf{Source}  \\
    \hline
    Satire&White House Chef Quits Because Trump Has Only Eaten Fast Food For 6 Months&halfwaypost\\
    Conspiracy&In Bizarre Response, Twitter Tells Trump It Does Not "Shadowban" While Admitting It Does&zerohedge\\
    Propaganda&CNN retracts story on investigation into Trump campaign adviser’s meeting with CEO of Russian fund&rt\\
    Neutral&Schiff disturbed by report that Trump not fully briefed on counters to Russian cyberattacks&upi\\
    Bias-Left&Remember Trump’s Promise Not to Touch Social Security? It’s Gone Now&nymag\\
    Bias-Right& Donald Trump Is America’s Julius Caesar&dailycaller\\
    \hline
    \end{tabular}
    \caption{Example snippets from articles about Donald Trump associated with source type.}
    \label{tab:my_label}
\end{table*}


 \begin{table*}[!htb]
    \begin{minipage}{.5\linewidth}
      \tiny
      \centering
      \begin{tabular}{llllll}
      \hline
      \textbf{Dataset} & \textbf{Type} &\textbf{Tokens-Body (Avg.)} &\textbf{Tokens-Title (Avg.)} & \textbf{Articles(\#)} & \textbf{Sources (\#)} \\
      \hline
      Community Interest & Satire & 287.65  & 10.48 & 1976 & 42 \\
      News Dataset& Conspiracy & 780.19 & 13.23 & 722 & 16\\
      & Propaganda & 752.13 & 10.88 & 1207 & 23\\
      & Neutral & 591.50 & 10.18 & 1365 & 40\\
      & Bias-Left & 700.73 & 11.42 & 1881 & 36 \\
      & Bias-Right & 557.88 & 11.44 &1284 & 23 \\
      \hline
      NELA 2019 & Satire & 233.37 & 10.85 & 1638 & 8\\
      & Conspiracy & 652.27 & 11.69 & 3710 & 16 \\
      & Propaganda & 538.69 & 11.37  & 6091 & 27\\
      & Neutral & 628.63 & 9.38 & 2366 & 11\\
      & Bias-Left & 670.09 &11.17  & 5262 & 23 \\
      & Bias-Right & 486.93 & 10.32 & 3857 & 17 \\
        \hline
  \end{tabular}
  \caption{Dataset statistics of community interest news dataset (CIND) and NELA 2019.}
  \label{dataset_statistics}
    \end{minipage}%
    \begin{minipage}{.65\linewidth}
    \small
      \centering
     \begin{tabular}{l|l}
    \hline
          \textbf{Train} & \#\\
    \hline
          NFS & 14685\\
          UFS & 2403\\
          CFS & 5413\\
    \hline
          \textbf{Test} & \#\\
    \hline
          NFS & 731\\
          UFS & 63\\
          CFS & 238\\
    \hline
     \end{tabular}
     \caption{Data statistics of ClaimBuster dataset.}
       \label{claimbuster:dataset_statistics}
    \end{minipage} 
    \vspace{-0.5cm}
\end{table*}

We categorized the sources of news articles in both datasets based on the false news taxonomy proposed by \citet{ZannettouSBK19}. The news types we selected from the taxonomy are satire, conspiracy, propaganda, biased (left \& right) as false news types and additionally neutral news as most credible news. For identifying the news sources in each category, we leveraged Media Bias Fact Check (MBFC) \footnote{https://mediabiasfactcheck.com/} which is an independent organization that manually annotates factuality and political leaning of media sources. Labels provided by MBFC have been widely used by the research community (e.g \citet{BalyKAGN18,NorregaardHA19,Barron-CedenoJM19}). We scraped MBFC labels\footnote{retrieved on 29.07.2020} and augmented the list with satire sources from the  \texttt{r/satire} subreddit\footnote{We marked the sources if their about/home pages explicitly indicate that they are satiric}. We explain the sources below with the traits of online information \citep{wardle2017information,ZannettouSBK19}.\\
\textbf{Satire} sources use irony, exaggeration and humour. Satiric articles do not aim to deceive the news consumer, but to entertain. However, if satire is taken seriously, it misinforms. We used the sources in the the \texttt{r/satire} subreddit and excluded sources that are not mutually exclusive (e.g https://www.newyorker.com/humor is also a bias source). \\
\textbf{Conspiracy} sources are not credible and mostly consist of articles that are not verifiable. These sources fabricate content with the intention to disinform. We extracted conspiracy sources from the conspiracy-pseudoscience category of MBFC. \\
\textbf{Propaganda} sources influence the news consumer in favor of a particular agenda. They may mislead in order to frame issues or individuals. We manually checked the questionable source category of MBFC to identify propaganda sources. MBFC provides tags to inform the reason why the source is questionable. One such tag is propaganda. Thus, we removed the sources that have labels other than propaganda in this tag.\\
\textbf{Neutral} sources are the most credible. They are least biased and their reporting is factual and verifiable.
We extracted least biased sources from MBFC as neutral sources. \\
\textbf{Biased} sources are strongly biased toward one ideology (typically: conservative/liberal) in their story selection and framing. We extracted biased sources from MBFC as bias sources. The bias category of the dataset contains 61\% of left and 39\% of bias-right sources. \\
After identifying source types of each article in both dataset, we removed sources that have samples of less than 10 articles and down-sampled sources that have more than 250 documents. Additionally, for CIND dataset, we removed the outliers for each category by computing the length of tokens of article bodies and by performing the local outlier factor algorithm \citep{BreunigKNS00} to yield its final dataset. The details of the CIND and NELA 2019 are shown in Table~\ref{dataset_statistics} and example articles are shown for each source type from CIND in Table \ref{tab:my_label}.

\vspace{-0.2cm}
\subsection{Dataset for Check-worthy Claim Detection \label{datasets:claim_buster}}
 To evaluate the model in claim-level, we used the ClaimBuster dataset \cite{ArslanHLT20} . It contains human annotated 23k short statements with a metadata, from all U.S presidential debates between 1960-2016. The dataset has been used for developing ClaimBuster. As part of the annotation process, \changemarker{the authors asked the coders to label the sentences as check-worthy factual claims (CFS) if they contain factual claims about which the public will be interested in learning about their veracity. Similarly, if the sentences contain factual statements but aren’t worth being fact-checked, they are annotated as unimportant factual sentences (UFS). Lastly, the coders considered the subjective sentences as non-factual sentences (NFS), such as opinions.}
 
\subsection{CSQA Dataset\label{commonsense}}
We used CSQA dataset  \citep{TalmorHLB19} for the model to learn commonsense knowledge. The dataset has been created based on the commonsense knowledge encoded in ConceptNET \citep{speer2017conceptnet}. The dataset contains 12k multiple-choice questions that have four choices and one correct answer.    

\begin{table*}[h!]
\centering
\small
    \begin{tabular}{lllllllll}
            \hline
    \rowcolor{white}
            \textbf{Feature} & \textbf{Model} & \textbf{Satire} & \textbf{Conspiracy} & \textbf{Propaganda} & \textbf{Neutral} & \textbf{Bias-Left} & \textbf{Bias-Right} & \textbf{F1-Macro} \\\hline
    
    \multirow{3}{*}{Title} & SVM & 45.91  & 21.14  & 32.34  &  37.29 & 40.56  & 15.04  & 32.05   \\
     & BERT & 54.38 & 33.33 & \textbf{46.67} & 46.30 & 49.37 &  19.24  & 41.40 \\
     & MTBERT &  \textbf{61.79} &  \textbf{33.78} & 45.71 &  \textbf{46.63} & \textbf{52.73} & \textbf{22.33}  & \textbf{43.83} \\ \hline
     \multirow{3}{*}{Body} & SVM  & \textbf{87.08} & 49.56 & 77.68 & 77.42   & 73.40 & 39.13 & 67.38  \\
      & HAN & 48.69  & 34.71  & 74.26 & 70.91 & 58.20  & 24.82 & 51.93 \\
    
    & HAN* & 67.41  & 32.17  & 66.67 & 60.23 & 56.24  & 28.32  & 51.84 \\
    & BERT  & 77.18 & 51.80 & 82.38 & \textbf{79.74}  & 72.00 & \textbf{40.99} &  66.90\\
    & MTBERT & 73.50 & \textbf{59.20}  & \textbf{83.22}  & 78.56 &  \textbf{73.54} & 37.17 &  \textbf{67.53} \\ \hline
    \multirow{3}{*}{Merged} & RDEL  & 75.72 & 45.53 & 55.02 & 63.11 & 59.28 & 26.74 & 54.66 \\
    & 3HAN & 45.08 & 25.86  & 52.25 & 46.84 & 34.62  & 18.15 & 39.81 \\
    & 3HAN* & 46.21  & 33.58  & 60.17  & 64.77 & 57.76  & 25.51 & 51.61 \\
    & BERT & \textbf{77.79}  & 60.87   & 83.15 & \textbf{85.66}  & 75.39  & \textbf{43.98} & \textbf{71.14} \\
    & MTBERT & 67.95 & \textbf{64.29} &  \textbf{86.09} & 74.11  & \textbf{76.60}  & 42.21  &  68.54 \\ \hline
    \end{tabular}
    \caption{Results for the forecasting experiment for false news classification. * indicates that the model using Numberbatch as embeddings. The bold scores are the best results per feature.}
    \label{forecasting:results}
\end{table*}

\section{Experiments and Evaluation \label{results}}
In our experiments, we investigate whether CSQA task helps  on (1) robustness on new events and changes of style by news publishers (Section \ref{predicting_future_section}), (2) classifying news from previously unseen publishers (Section \ref{predicting_source_type_unseen_article_from_unseen_source}), and (3) discovering check-worthy claims (Section \ref{check-worthy_detection}). We report per-class and average macro F1 scores for each experiment.


\subsection{Robustness on New Events and Changes of Style by News Publishers \label{predicting_future_section}}
In this analysis, we test the models' robustness against new events or the style changes by the news publishers. For this, we adopt the forecasting experiment proposed by \citet{BozarthB20}, first we extract sources that published articles in between 2015 and 2019 from the CIND dataset. We use data samples whose published year is earlier than 2019 as training and the rest is as test set.

\begin{figure*}[t]
    \centering
    \includegraphics[width=0.9\textwidth]{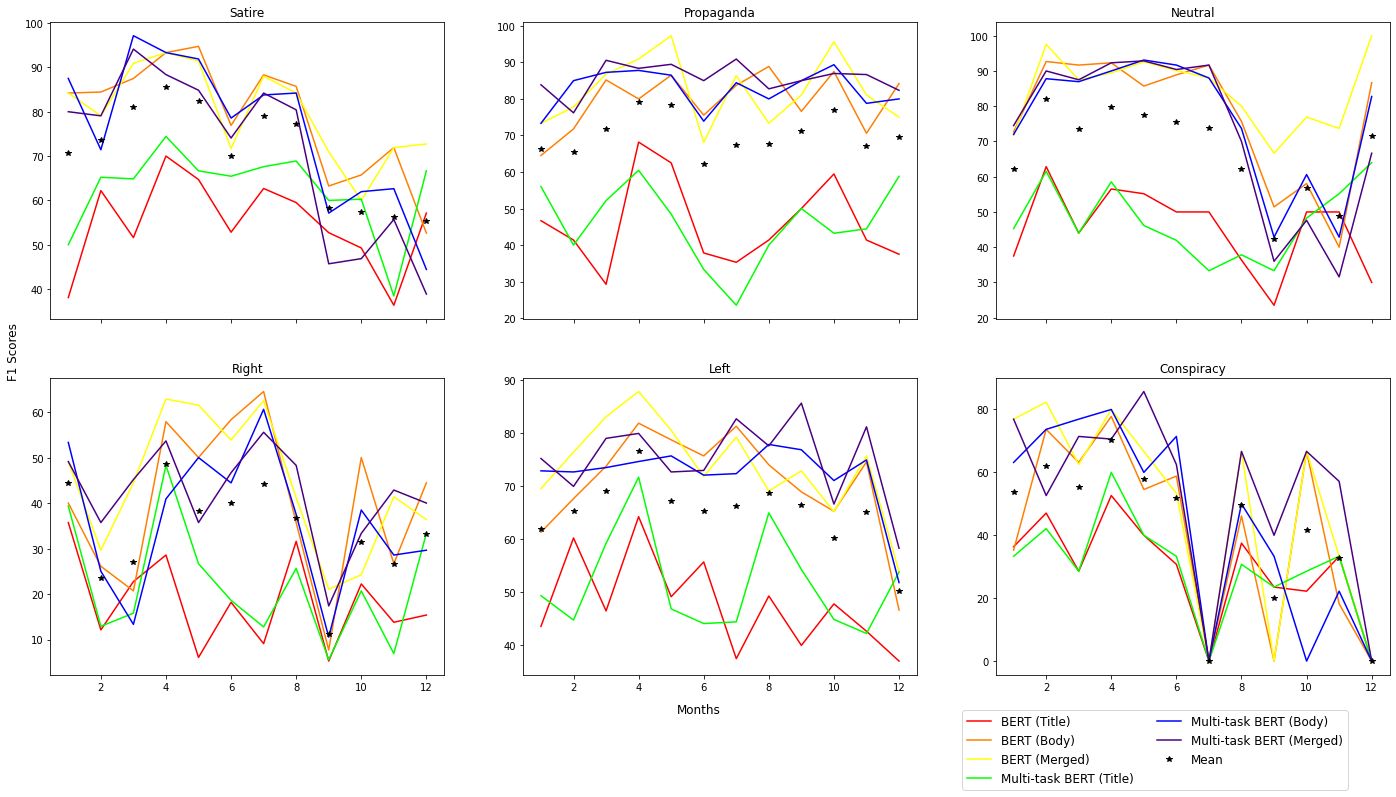}
    \caption{Comparison of single task BERT and MTBERT in forecasting task on CIND dataset. X-axis indicates the months in 2019, y-axis indicates the F1 macro scores.}
    \label{fig:forecasting_BERT}
\end{figure*}

Table \ref{forecasting:results} shows the macro-F1 scores of the models for each category and the Figure \ref{fig:forecasting_BERT} shows the macro-F1 scores of BERT variations for each month in 2019. Overall, BERT models outperform SVMs, HANs and RDEL in this task, however, detecting right-lean articles are hard for all of the models. Additionally, even though MTBERT using merged features cannot outperform the single task in terms of macro-F1, it \changemarker{has performance gains in the classes conspiracy (3.42\%), propaganda (2.87\%) and left (1.21\%).} \changemarker{As shown in Figure \ref{fig:forecasting_BERT}, the performances of BERT variants in the satire class degrade but the performances in propaganda and left types are stable throughout the year. Furthermore, the fluctuations in MBERT (with the merged features) are less than the single task BERT on propaganda across the year, which could be more generalized at identifying propaganda throughout a year.} 

\begin{figure*}[t!]
    \centering
    \begin{subfigure}[b]{\textwidth}
        \centering
        \includegraphics[width=1.0\linewidth]{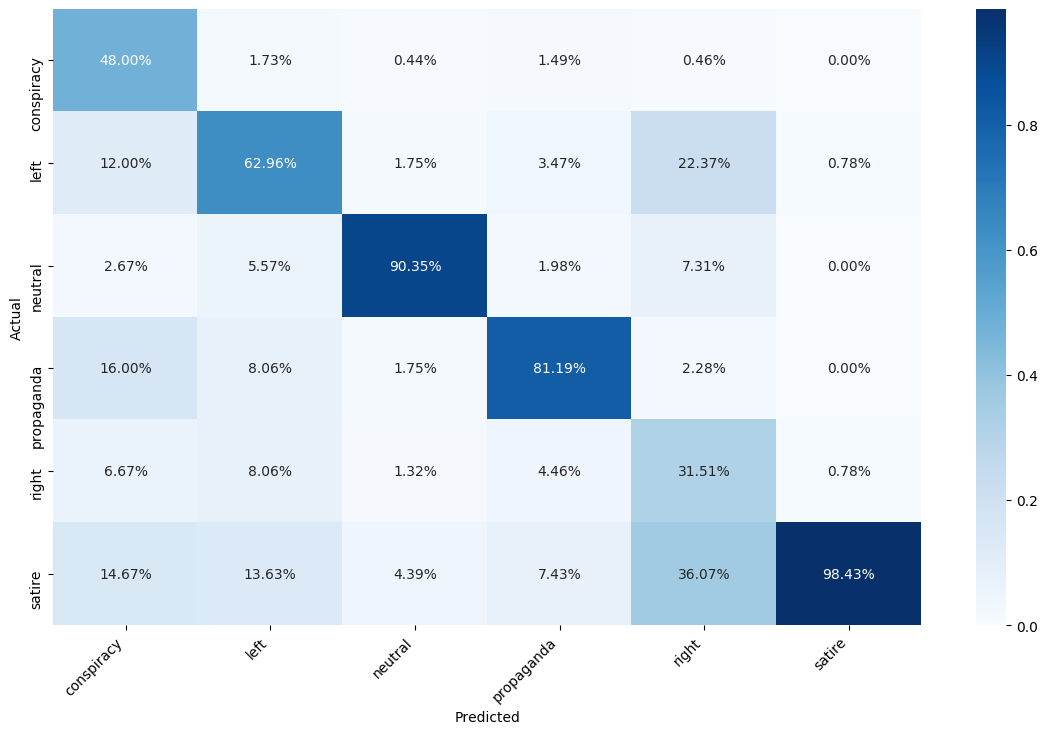}
        \caption{Single BERT}
    \end{subfigure}%
    \\ 
    \begin{subfigure}[b]{\textwidth}
        \centering
        \includegraphics[width=1.0\linewidth]{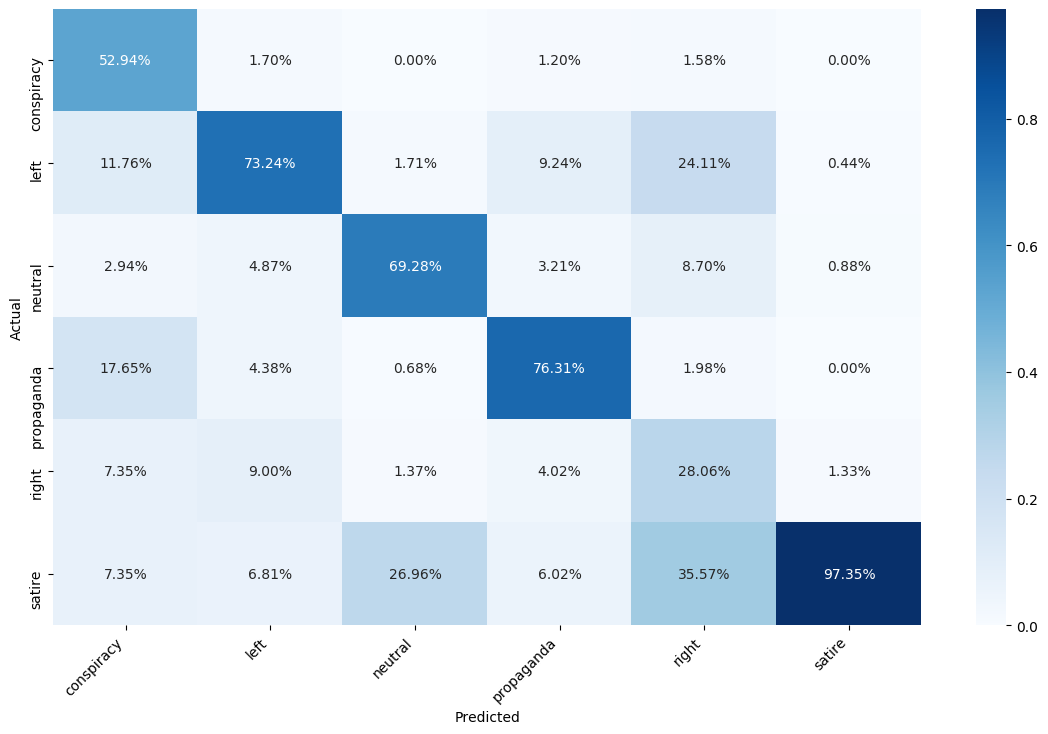}
        \caption{Multitask BERT}
    \end{subfigure}
    \caption{Confusion Matrices of Predictions from Single BERT and Multitask BERT on the forecasting task.}
    \label{cm:forecasting}
\end{figure*}

\changemarker{Moreover, as shown in Figure~\ref{cm:forecasting}, MTBERT produces higher number of true positive samples than the single task, BERT in conspiracy articles. However, the single task BERT is better at identifying neutral articles for the forecasting task.}

\subsection{Classifying News from Previously Unseen Publishers\label{predicting_source_type_unseen_article_from_unseen_source}}
To meet the condition of unseen publishers, we organize our tests such that the publisher of a news article has not been encountered before. To see how MTBERT can generalize in such a scenario, we adopt the evaluation scheme for predicting unseen sources \citep{BozarthB20} and apply it to NELA and CIND datasets. First, we group news articles based on source (reuters, fox news, etc.) under each source type (conspiracy, propaganda and so on). From each source type we randomly sample 90\% sources as training set and the rest as test set and repeat it 5 times. We report the mean and standard deviation of macro-F1 scores of sets in Table \ref{outdomain:results}.

Similar to forecasting experiments, both BERT-based models outperform all baseline models, although a significant performance drop is observed in all models compared to forecasting experiments. This is expected since in forecasting experiments, test set contains articles from sources that may also have some samples in training set. It makes the predictions in forecasting easier for the models considering that they can also learn from stylistic features of sources. We observe that satire sources are the best detected types in the CIND dataset, but not in the NELA dataset, because the test sets contain only one satire source and adds a new challenge for the models. 

\changemarker{We additionally ran a paired t-test on the scores. MTBERT combining all features significantly outperformed the single task version on the CIND (at p value of 0.001, with a large effect size: cohen's d 2.339). Yet single task BERT achieves 0.39\% better F1 score than the MTBERT on NELA dataset, and no significant difference is observed. The reason could be the difference between the construction of the datasets. For example, CIND contains selected samples by Reddit users, which might have clickbait/check-worthy statements. However, NELA samples are collected from RSS of websites where publisher preference has a significant role. That also explains why a title of an article does not improve the results for the NELA.} 

\begin{figure*}[t!]
    \centering
    \begin{subfigure}[b]{\textwidth}
        \centering
        \includegraphics[width=1.0\linewidth]{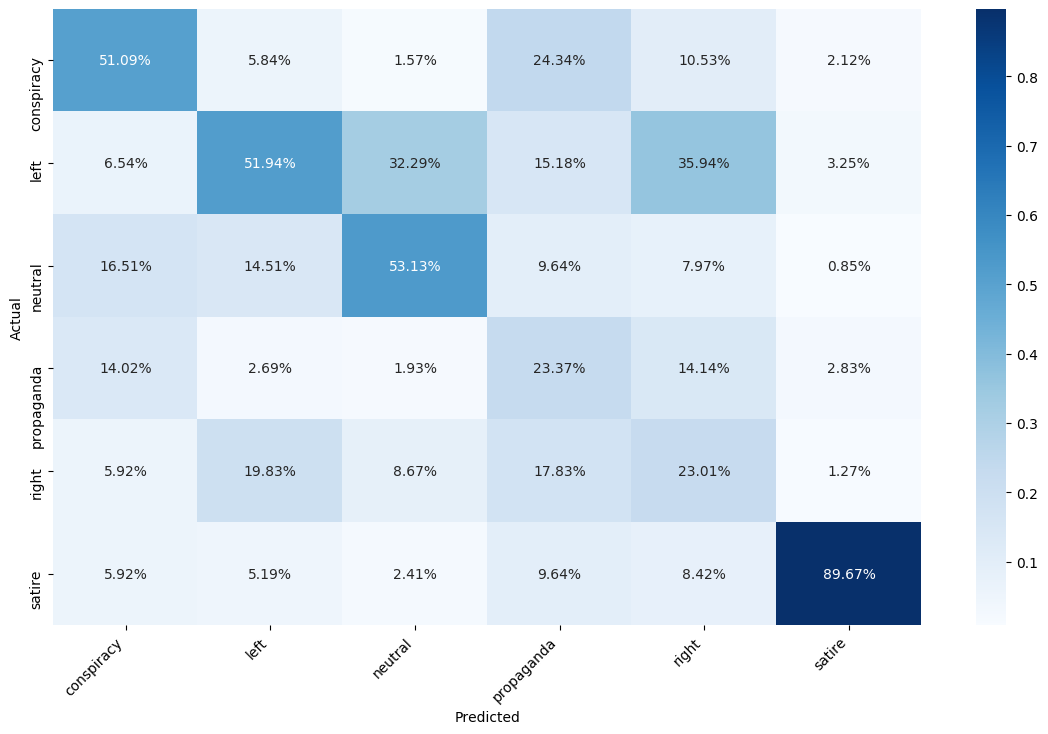}
        \caption{Single BERT}
    \end{subfigure}%
    \\ 
    \begin{subfigure}[b]{\textwidth}
        \centering
        \includegraphics[width=1.0\linewidth]{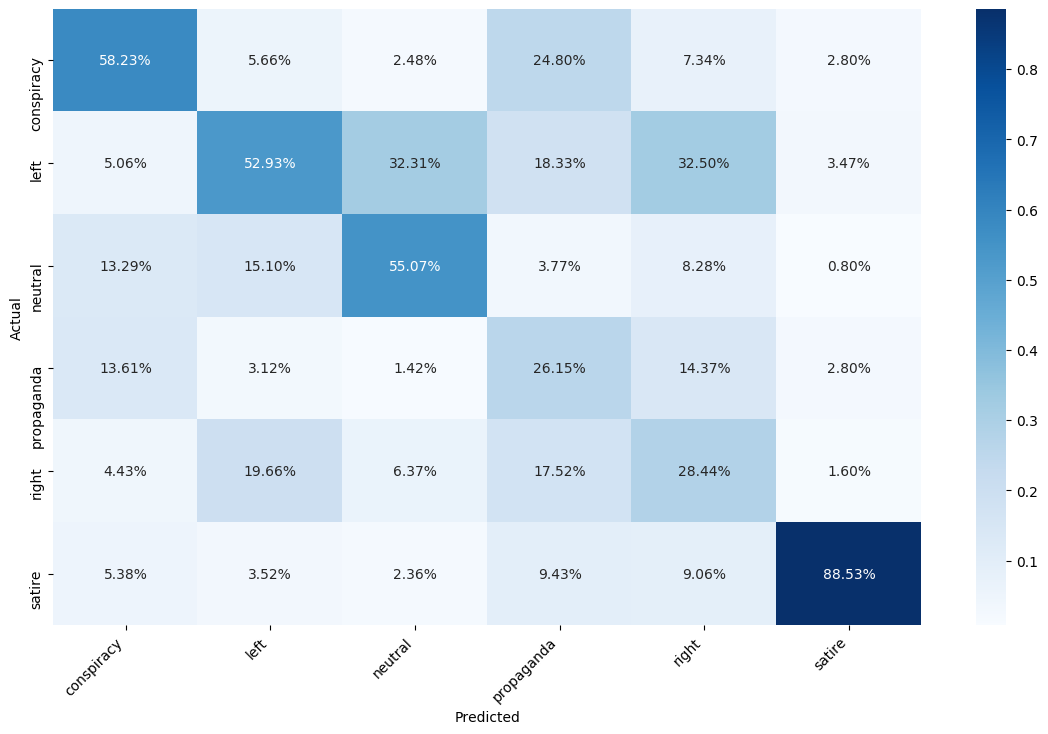}
        \caption{Multitask BERT}
    \end{subfigure}
    \caption{Confusion Matrices of Predictions of Single BERT and Multitask BERT from the unseen prediction task on CIND dataset.}
    \label{cm:unseen_prediction}
\end{figure*}

\begin{figure*}[t!]
    \centering
    \begin{subfigure}[b]{\textwidth}
        \centering
        \includegraphics[width=1.0\linewidth]{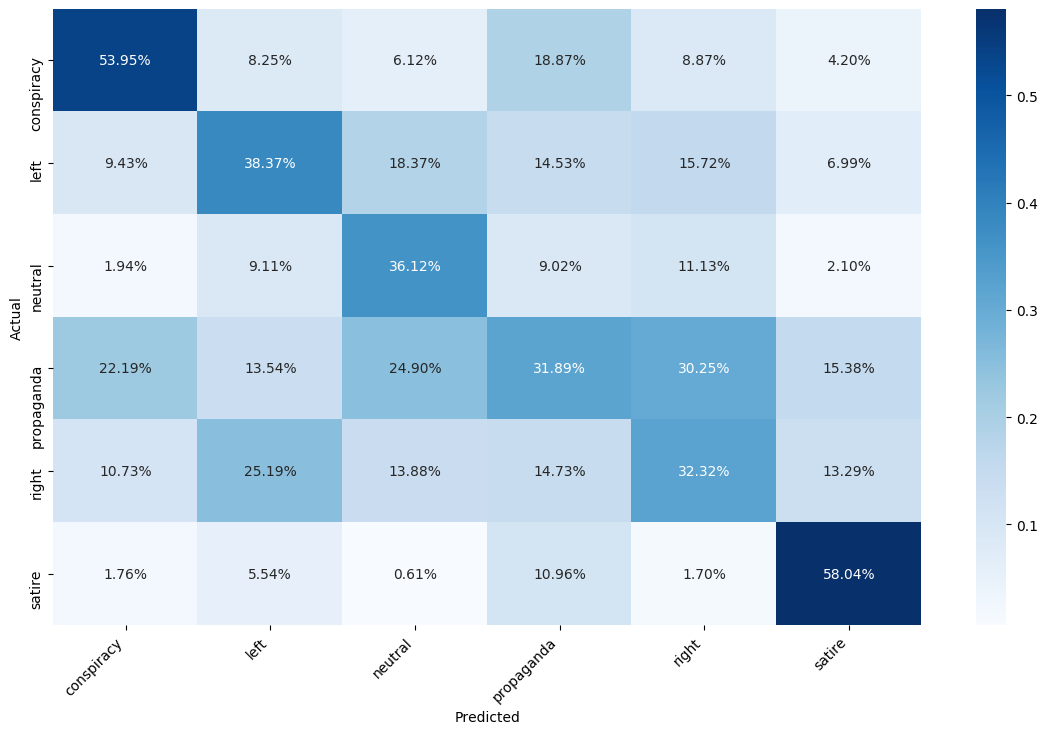}
        \caption{Single BERT}
    \end{subfigure}%
    \\ 
    \begin{subfigure}[b]{\textwidth}
        \centering
        \includegraphics[width=1.0\linewidth]{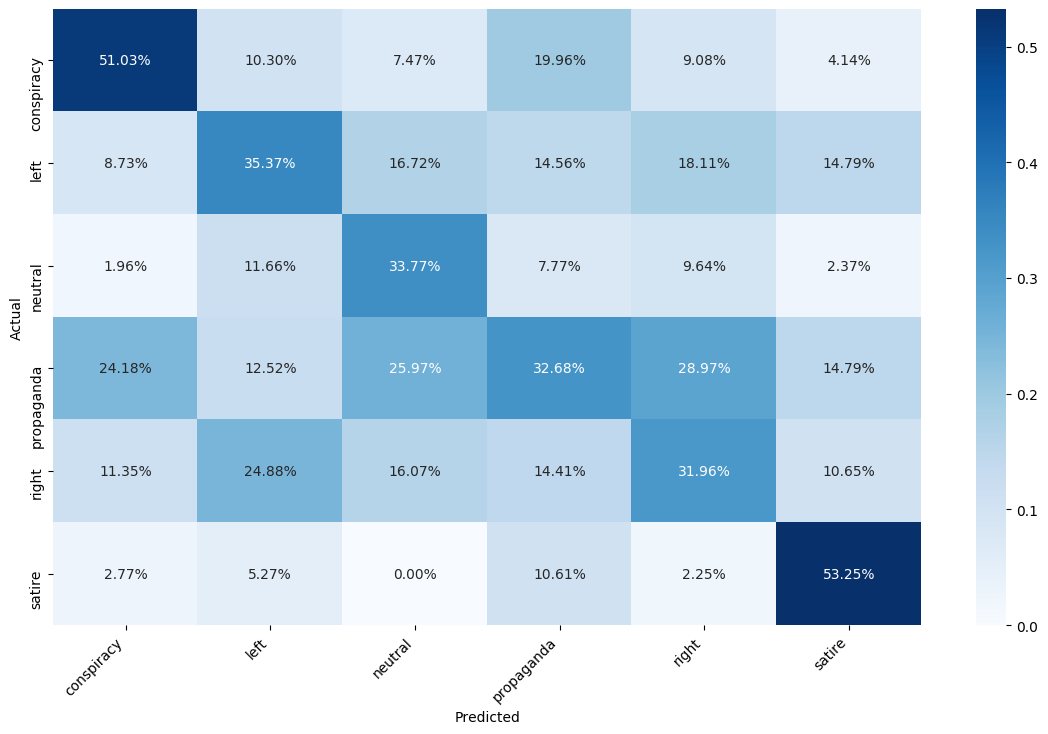}
        \caption{Multitask BERT}
    \end{subfigure}
    \caption{Confusion Matrices of Predictions from Single BERT and Multitask BERT on NELA dataset.}
    \label{cm:kfold}
\end{figure*}

\changemarker{Like the forecasting task, commonsense knowledge has a positive effect on true positive conspiracy samples on CIND, as seen in Figure~\ref{cm:unseen_prediction}. However, both BERT models misclassified the samples from propaganda sources mostly as right-lean article (shown in Figures~\ref{cm:unseen_prediction},\ref{cm:kfold}).}

\begin{table*}[h!]
\centering
\small
        \begin{tabular}{llllllllll}
            \hline
            \textbf{Dataset} & \textbf{Feature} & \textbf{Model} & \textbf{Satire} & \textbf{Conspiracy} & \textbf{Propaganda} & \textbf{Neutral} & \textbf{Bias-Left} & \textbf{Bias-Right} & \textbf{F1-Macro} \\\hline
            \multirow{13}{*}{CIND} & \multirow{3}{*}{Title} & SVM & \multicolumn{1}{l}{42.68} &                \multicolumn{1}{l}{23.78} & \multicolumn{1}{l}{20.19} &  \multicolumn{1}{l}{22.07} & \multicolumn{1}{l}{47.28 } & \multicolumn{1}{l}{12.47} & \multicolumn{1}{l}{28.08} \\
            & &BERT &\multicolumn{1}{l}{53.47} & \multicolumn{1}{l}{\textbf{37.20}} & \multicolumn{1}{l}{26.02} & \multicolumn{1}{l}{\textbf{30.75}}  & \multicolumn{1}{l}{50.24} & \multicolumn{1}{l}{\textbf{17.47}} & \multicolumn{1}{l}{35.86} \\
            & &MTBERT & \multicolumn{1}{l}{\textbf{53.50}}  & \multicolumn{1}{l}{35.68} & \multicolumn{1}{l}{\textbf{27.31}} & \multicolumn{1}{l}{29.90} & \multicolumn{1}{l}{\textbf{52.11}} & \multicolumn{1}{l}{17.27} & \multicolumn{1}{l}{\textbf{35.96}} \\\cline{2-10}
            &\multirow{6}{*}{Body}& SVM & \multicolumn{1}{l}{72.30}  & \multicolumn{1}{l}{\textbf{44.90}} & \multicolumn{1}{l}{\textbf{42.75}} &  \multicolumn{1}{l}{39.27}  & \multicolumn{1}{l}{\textbf{53.62}}  & \multicolumn{1}{l}{19.77}  & \multicolumn{1}{l}{42.75} \\

            && HAN & \multicolumn{1}{l}{61.18} & \multicolumn{1}{l}{39.34} & \multicolumn{1}{l}{23.36} &  \multicolumn{1}{l}{34.05} &  \multicolumn{1}{l}{44.41} &  \multicolumn{1}{l}{18.98} & \multicolumn{1}{l}{36.89}\\

            && HAN*  & \multicolumn{1}{l}{60.15}   & \multicolumn{1}{l}{32.24} & \multicolumn{1}{l}{19.42}  & \multicolumn{1}{l}{31.26} & \multicolumn{1}{l}{46.33} & \multicolumn{1}{l}{17.15} & \multicolumn{1}{l}{34.43} \\
                  
            && BERT  & \multicolumn{1}{l}{\textbf{78.82}} & \multicolumn{1}{l}{42.63} & \multicolumn{1}{l}{23.00}  & \multicolumn{1}{l}{38.94} & \multicolumn{1}{l}{51.27}  & \multicolumn{1}{l}{\textbf{26.57}}  & \multicolumn{1}{l}{43.54} \\
         && MTBERT  & \multicolumn{1}{l}{75.09}  & \multicolumn{1}{l}{36.69} & \multicolumn{1}{l}{23.6} &  \multicolumn{1}{l}{\textbf{49.78}} & \multicolumn{1}{l}{51.41}  & \multicolumn{1}{l}{25.64} & \multicolumn{1}{l}{\textbf{43.70}} \\ \cline{2-10}

        &\multirow{6}{*}{Merged}& RDEL  &  \multicolumn{1}{l}{70.11} &  \multicolumn{1}{l}{37.88} &  \multicolumn{1}{l}{25.12} &  \multicolumn{1}{l}{30.37} &  \multicolumn{1}{l}{48.99} &  \multicolumn{1}{l}{21.76} &  \multicolumn{1}{l}{39.04}  \\

         && 3HAN & \multicolumn{1}{l}{54.36} & \multicolumn{1}{l}{29.37}  &  \multicolumn{1}{l}{20.88}  & \multicolumn{1}{l}{31.44} & \multicolumn{1}{l}{44.83}  & \multicolumn{1}{l}{17.20}  & \multicolumn{1}{l}{33.01}  \\

         && 3HAN* & \multicolumn{1}{l}{51.57}  & \multicolumn{1}{l}{31.05} & \multicolumn{1}{l}{19.85}  & \multicolumn{1}{l}{34.96} & \multicolumn{1}{l}{42.62} &  \multicolumn{1}{l}{16.94}  & \multicolumn{1}{l}{32.83} \\

         && BERT & \multicolumn{1}{l}{76.46} &  \multicolumn{1}{l}{43.05} & \multicolumn{1}{l}{23.34} & \multicolumn{1}{l}{46.90} & \multicolumn{1}{l}{51.67} & \multicolumn{1}{l}{23.35} &  \multicolumn{1}{l}{44.13} \\
        && MTBERT &  \multicolumn{1}{l}{\textbf{78.55}} &  \multicolumn{1}{l}{\textbf{46.95}} &  \multicolumn{1}{l}{\textbf{25.67}} &  \multicolumn{1}{l}{\textbf{48.01}} &  \multicolumn{1}{l}{\textbf{53.95}}    &  \multicolumn{1}{l}{\textbf{27.89}} &  \multicolumn{1}{l}{\textbf{46.83}} \\

        \hline
        \multirow{13}{*}{NELA} & \multirow{3}{*}{Title} & SVM & \multicolumn{1}{l}{4.08} & \multicolumn{1}{l}{29.51} & \multicolumn{1}{l}{\textbf{33.30}} & \multicolumn{1}{l}{12.78} & \multicolumn{1}{l}{29.23}  & \multicolumn{1}{l}{19.59} &  \multicolumn{1}{l}{21.42} \\
        
        && BERT & \multicolumn{1}{l}{4.59}  & \multicolumn{1}{l}{38.50} & \multicolumn{1}{l}{27.94} & \multicolumn{1}{l}{17.89} & \multicolumn{1}{l}{28.32} & \multicolumn{1}{l}{20.91} &  \multicolumn{1}{l}{23.02} \\
            & &MTBERT & \multicolumn{1}{l}{\textbf{6.49}} & \multicolumn{1}{l}{\textbf{39.80}} & \multicolumn{1}{l}{28.71} &  \multicolumn{1}{l}{15.55} & \multicolumn{1}{l}{\textbf{30.16}} & \multicolumn{1}{l}{\textbf{21.22}} & \multicolumn{1}{l}{\textbf{23.65}} \\ \cline{2-10}
        &\multirow{6}{*}{Body}& SVM & \multicolumn{1}{l}{8.67} & \multicolumn{1}{l}{43.03} &  \multicolumn{1}{l}{35.18} & \multicolumn{1}{l}{\textbf{22.28}}   &  \multicolumn{1}{l}{36.02} & \multicolumn{1}{l}{26.47} & \multicolumn{1}{l}{28.61} \\
        && HAN & \multicolumn{1}{l}{3.62} & \multicolumn{1}{l}{38.43} & \multicolumn{1}{l}{36.12}  &  \multicolumn{1}{l}{20.99} &  \multicolumn{1}{l}{29.41}  &  \multicolumn{1}{l}{29.56}  & \multicolumn{1}{l}{26.35} \\
        &&  HAN* & \multicolumn{1}{l}{10.25}  & \multicolumn{1}{l}{35.41} & \multicolumn{1}{l}{33.04} & \multicolumn{1}{l}{19.12} & \multicolumn{1}{l}{30.72} & \multicolumn{1}{l}{26.73} & \multicolumn{1}{l}{25.88} \\  
         && BERT &\multicolumn{1}{l}{\textbf{16.28}} & \multicolumn{1}{l}{44.08} & \multicolumn{1}{l}{\textbf{38.74}}  & \multicolumn{1}{l}{17.25} & \multicolumn{1}{l}{\textbf{38.55}}  & \multicolumn{1}{l}{34.44}
         & \multicolumn{1}{l}{\textbf{31.56}} \\
         && MTBERT & 10.27 & \textbf{48.24} & 38.45  &  15.12 &  37.4 & \textbf{36.93}  & 31.07 \\ \cline{2-10}
         &\multirow{6}{*}{Merged}& RDEL & \multicolumn{1}{l}{11.52} &  \multicolumn{1}{l}{39.87} &  \multicolumn{1}{l}{35.12} &  \multicolumn{1}{l}{\textbf{21.79}} &  \multicolumn{1}{l}{33.43} &  \multicolumn{1}{l}{28.80}  &  \multicolumn{1}{l}{28.42}  \\
         && 3HAN & \multicolumn{1}{l}{9.79} & \multicolumn{1}{l}{37.79} & \multicolumn{1}{l}{35.92} &  \multicolumn{1}{l}{18.59} & \multicolumn{1}{l}{30.98} & \multicolumn{1}{l}{27.81}  & \multicolumn{1}{l}{26.86} \\
         && 3HAN* &  \multicolumn{1}{l}{16.57} &  \multicolumn{1}{l}{36.27} &  \multicolumn{1}{l}{32.90}  &   \multicolumn{1}{l}{20.85} &  \multicolumn{1}{l}{32.20}  &  \multicolumn{1}{l}{27.53}  &  \multicolumn{1}{l}{27.72} \\
         & & BERT & \multicolumn{1}{l}{13.92} & \multicolumn{1}{l}{\textbf{50.04}} & \multicolumn{1}{l}{\textbf{37.94}} & \multicolumn{1}{l}{20.48} & \multicolumn{1}{l}{\textbf{38.33}} & \multicolumn{1}{l}{\textbf{31.24}} & \multicolumn{1}{l}{\textbf{31.99}} \\
        && MTBERT &   \multicolumn{1}{l}{\textbf{17.63}} &  \multicolumn{1}{l}{46.29} &  \multicolumn{1}{l}{37.89} &  \multicolumn{1}{l}{21.12} &  \multicolumn{1}{l}{36.47}    &  \multicolumn{1}{l}{30.22} &  \multicolumn{1}{l}{31.60} \\\hline
        \end{tabular}
    \caption{Results of the experiment that the news articles in test set are from previously unseen source. The cells indicate the mean and standard deviation of 5 folds F1 scores for each class and * indicates that the model using Numberbatch as embeddings. The bold scores are the best results per feature and dataset.}
    \vspace{-0.5 cm}
    \label{outdomain:results}
\end{table*}

\subsection{Evaluating Commonsense Knowledge for Identifying Check-worthy Claims\label{check-worthy_detection}}
Finally, we assess the help of commonsense knowledge in detecting check-worthy claims. Therefore, we evaluate the MTBERT for check-worthy claim detection task (see Section~\ref{datasets:claim_buster}). We use the splits provided by the authors \cite{ArslanHLT20}. 

\begin{figure*}[t!]
    \centering
    \begin{subfigure}[b]{\textwidth}
        \centering
        \includegraphics[width=1.0\linewidth]{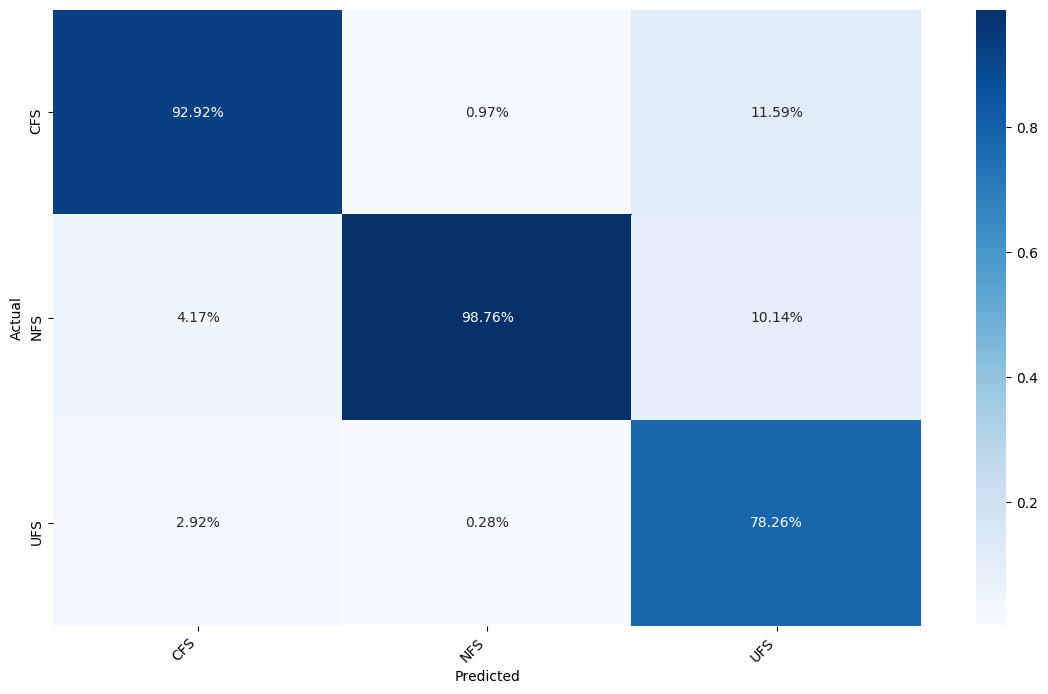}
        \caption{Single BERT}
    \end{subfigure}%
    \\ 
    \begin{subfigure}[b]{\textwidth}
        \centering
        \includegraphics[width=1.0\linewidth]{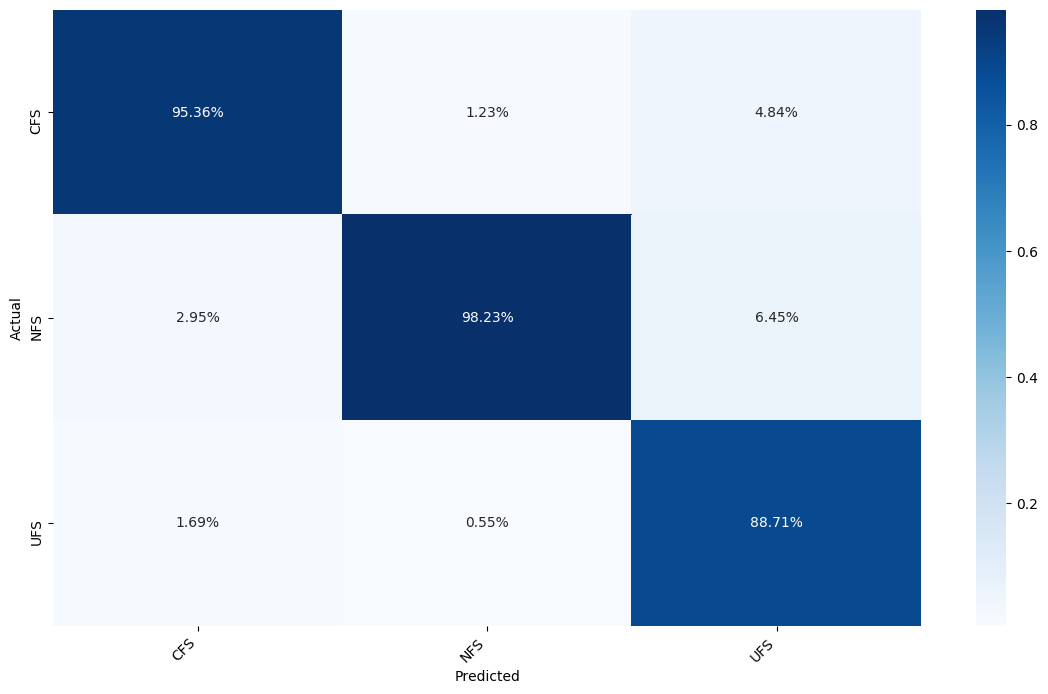}
        \caption{Multitask BERT}
    \end{subfigure}
    \caption{Confusion Matrices of Predictions from Single BERT and Multitask BERT on the check-worthiness detection task.}
    \label{cm:claim_check}
\end{figure*}

Table~\ref{tab:claim_buster_3} shows the performance of the models. Utilizing commonsense knowledge significantly improved the scores of each class in the task. \changemarker{As shown in Figure~\ref{cm:claim_check}, while the single task BERT confused with un/important sentences, MBERT could correctly classified some of the misclassified samples. It implies that commonsense knowledge could help detection of check-worthy claims from the all types of facts in the dataset.}

\changemarker{Adding CSQA as a downstream task in checkworthy claim detection was more effective on true positives than multiclass false news classification with the CSQA. That could be due to data formats of commonsense question answering task and check-worthy claim detection task are similar. It had a positive effect on knowledge transfer between the tasks. Both tasks contain short texts, one or two sentences, while samples in the datasets of false news detection tasks are composed of long texts that might bring about less overlapping between the tokens from the task inputs.}

\begin{table}[h!]
    \small
    \centering
    \begin{tabular}{llll}
    \hline
        \textbf{Model} & \textbf{CFS} & \textbf{NFS} & \textbf{UFS}\\
        \hline
        ClaimBuster API & 37.41 & 94.13 & 9.76 \\
        ClaimBuster & 83.96 & 94.72 & 47.31 \\
        \hline
        Single BERT & 93.31 & 98.21 & 81.82 \\
        MTBERT & \textbf{95.16} & \textbf{98.36} & \textbf{88.00} \\
        \hline
    \end{tabular}
    \caption{F1 scores of each class in check-worthy claim detection task. The bold scores are the best results.}
        \vspace{-0.5cm}
    \label{tab:claim_buster_3}
\end{table}

\section{Discussion}
\changemarker{In the experiments on the false news detection task, we found that detecting right-lean articles, in general, was not helpful among the classifiers. The reason for the difficulty in detecting right-lean articles could be a potential data bias. Social media and crowd-sourced platforms tend to lend high visibility to viral content, which includes right-leaning and left-leaning news publishers that are by definition extreme in their worldviews, coupled with a sensationalist tone. This issue also reflects the existing dataset NELA and our newly collected dataset CIND. In our study, we observed that right-leaning articles are mostly misclassified as left-leaning articles. The difficulty of detecting right-leaning articles is also observed by \citet{SteinPKBR18} for the hyperpartisan news task. Additionally, \citet{BozarthB20}~ analyzed that fake news are mostly misclassified as right-leaning news and observed that right-leaning news publishers sometimes campaigned false information. The significant difference of our study from the prior studies \citep{SteinPKBR18, BozarthB20} is that we tackle the problem as a multi-class news type classification because different fake news types may have different implications. Biased sources can also misinform the readers \citep{ZannettouSBK19,wardle2017information}, and fine-grained detection is vital for prioritizing what should be fact-checked. } \\
\changemarker{
We attempt to transfer commonsense knowledge to BERT representations implicitly with a multi-tasking approach. We achieved better performance on the check-worthy claim detection task due to the similar data type with CQSD. A sentence-based approach could be utilized to improve the performance in false news detection tasks and transfer knowledge more effectively than the current method. Also, explicit methods could be used for false news detection tasks. For example, a new task could be introduced, a plausibility detection task on news articles where annotators would evaluate the degree of believability on articles. And then, this task could be used as a downstream task.  
}


\section{Conclusion}
In this paper, we explore the impact of commonsense knowledge in the tasks of false news classification and check-worthy claim detection. To learn commonsense knowledge implicitly, we fine tune BERT jointly with CSQA for each tasks. The results show that the proposed model can improve the predictions \changemarker{of minority classes in the datasets (e.g conspiracy in CIND, CFS in check-worthy claim detection task). Also, similar formats of the inputs such as in CSQD and check-worthy claim detection could have a positive effect on performance.}\\
In conclusion, we introduced a new challenging dataset for a false news classification task and to our knowledge, this is the first work that examines the effects of using commonsense knowledge on false news classification and check-worthy claim detection tasks.

\bibliography{anthology,eacl2021}
\bibliographystyle{acl_natbib}

\onecolumn
\section*{Appendix}
\subsection{Details on CIND}
\subsubsection{Sample Distributions based on source type and subreddit}
\begin{figure*}
    \centering
    \begin{subfigure}[b]{\textwidth}
        \centering
        \includegraphics[width=1.0\linewidth]{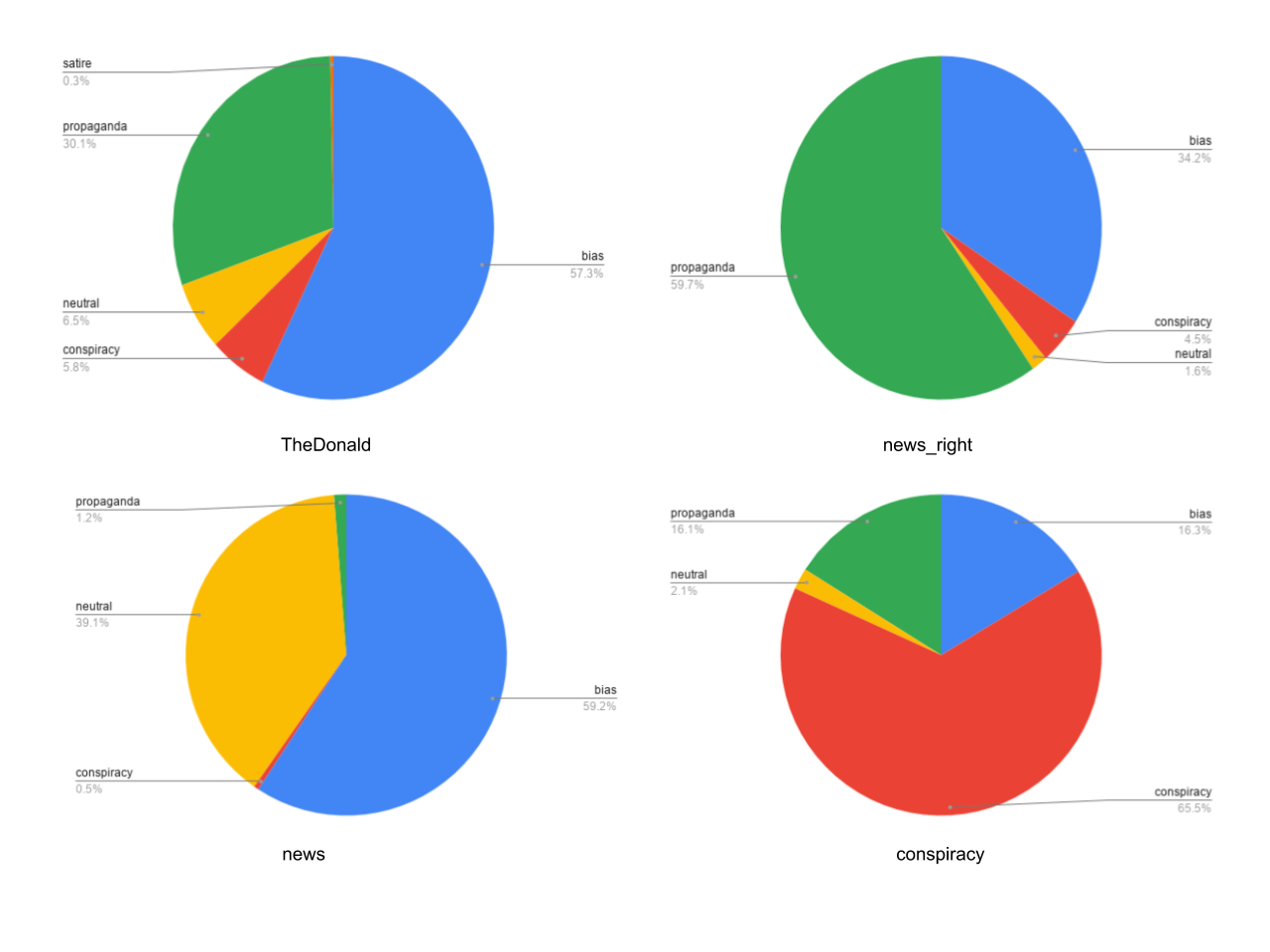}
    \end{subfigure}%
    \\ 
    \begin{subfigure}[b]{\textwidth}
        \centering
        \includegraphics[width=1.0\linewidth]{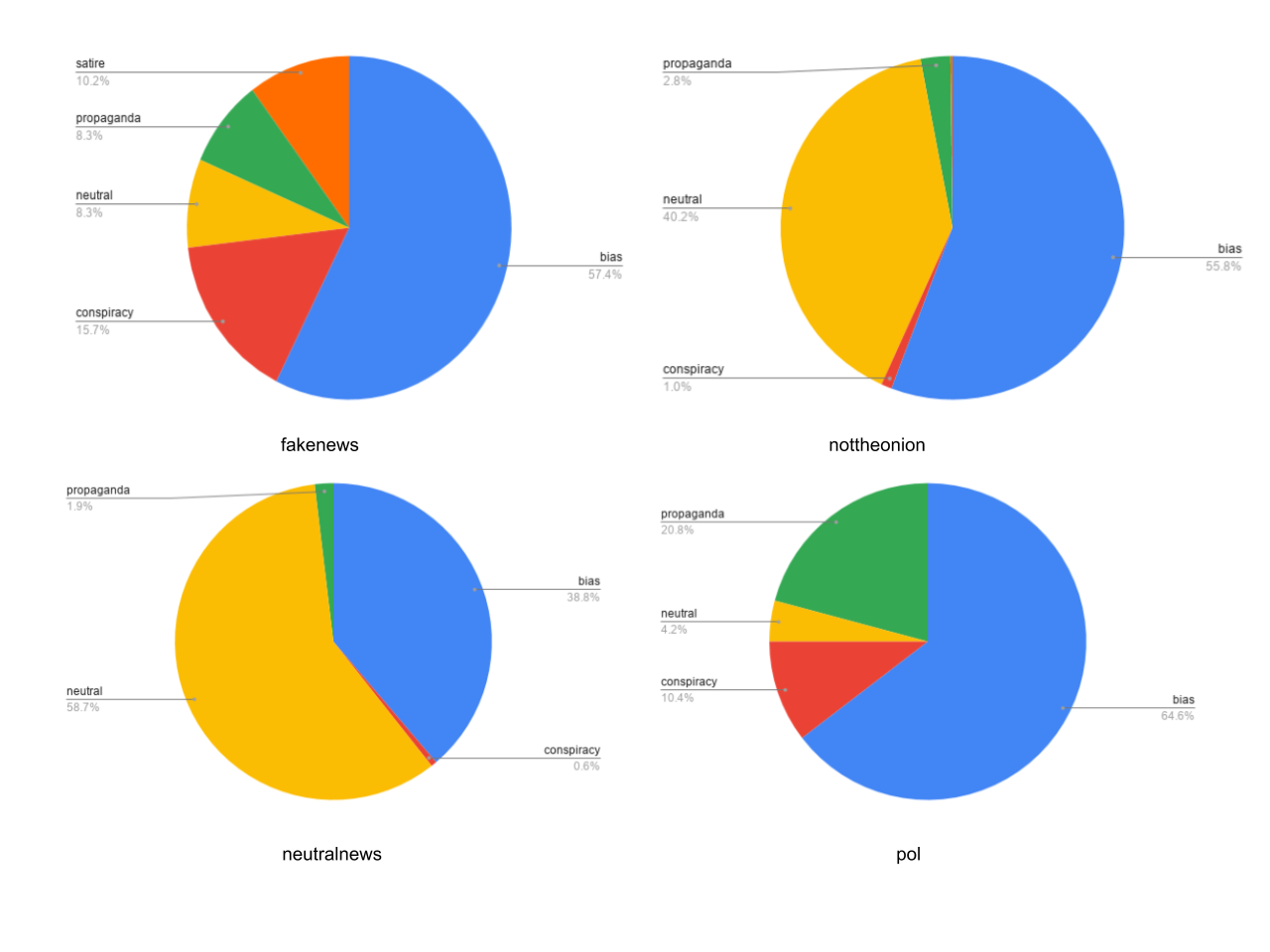}
    \end{subfigure}
    \caption{Sample distributions per source type and per subreddit.}
        \label{cind:distribution}
\end{figure*}

\begin{figure*}[t!]
    \begin{subfigure}[b]{\textwidth}
        \centering
        \includegraphics[width=1.0\linewidth]{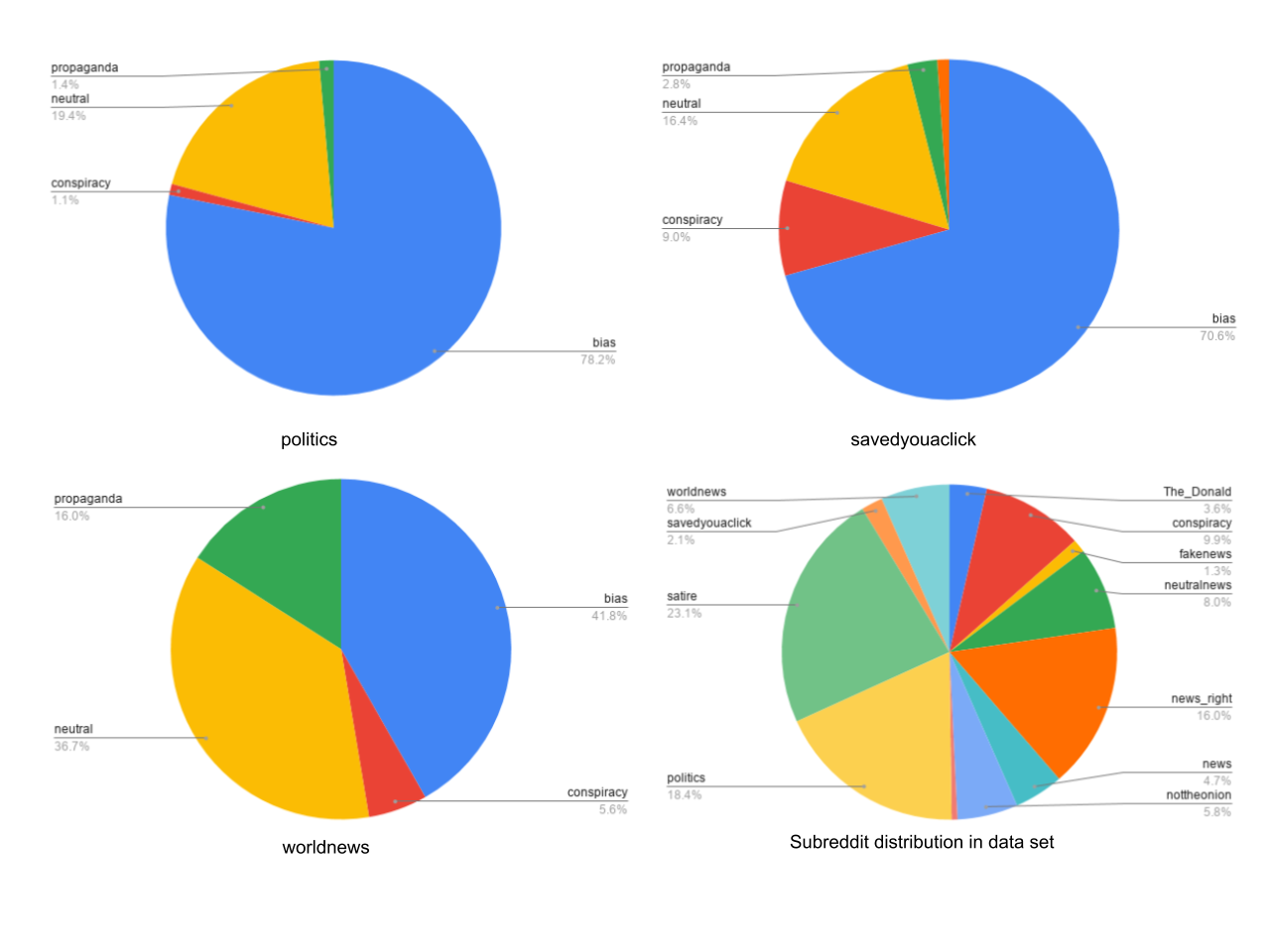}
    \end{subfigure}
    \caption{Sample distributions per source type and per subreddit. (Cont.)}
        \label{cind:distribution2}
\end{figure*}

\changemarker{This section gives more details about CIND. Figure~\ref{cind:distribution} and \ref{cind:distribution2} show the distributions of source types in each subreddit that we used. In the diagrams, \texttt{bias} tag combines bias-right and bias-left articles. Even though we could expect that some subreddits contain only specific source types, some source types are shared within multiple subreddits. For instance, neutral sources are seen in each subreddits.}

\subsubsection{Sample distributions on forecasting and unseen prediction tasks}
\begin{table}[!h]
     \begin{tabular}{lllllll}
    \hline
          \textbf{Split} & \textbf{Conspiracy} & \textbf{Bias-Left} & \textbf{Neutral} & \textbf{Propaganda} & \textbf{Bias-Right} & \textbf{Satire}\\
    \hline
          \textbf{Train} & 574 & 955 & 844 & 847 & 1019 & 859\\
          \textbf{Test} & 50 & 399 & 257 & 227 & 130 & 437 \\
    \hline
     \end{tabular}
     \caption{Data splits for the forecasting task.}
    \label{data_split:fc}
\end{table}

\begin{table}[!h]
     \begin{tabular}{llllllll}
    \hline
          \textbf{Fold} &\textbf{Split} & \textbf{Conspiracy} & \textbf{Bias-Left} & \textbf{Neutral} & \textbf{Propaganda} & \textbf{Bias-Right} & \textbf{Satire}\\
    \hline
    1 & Train&	662&	1610&	1146&	1102	&1087	&1899 \\
     & Test	&60	&271&	219	&105&	192	&77 \\
    2 & Train&	689	&1467&	1330&	1189&	1198&	1724 \\
    & Test	&33	&414&	35&	21	&81	&252 \\
    3 & Train&	493	&1500&	1301&	1167&	1226&	1864 \\
    & Test&	229	&381&	64&	40	&53	&112 \\
    4 & Train&	666	&1653&	1072&	1173&	1198&	1686 \\
     & Test&	56&	228	&293&	34&	81&	290\\
    5 & Train&	648&	1770&	1162&	1094&	1057&	1859 \\
     & Test &	74&	111&	203&	113	&222&	117\\
    \hline
     \end{tabular}
     \caption{Data splits for the unseen prediction task.}
    \label{data_split:unseen_prediction}
\end{table}

\changemarker{Tables~\ref{data_split:fc}} and \ref{data_split:unseen_prediction} include overview of the class distributions for the forecasting and unseen prediction tasks on CIND.

\subsection{Experimental Settings}
\subsubsection{Preprocessing}
For false news detection task, we apply the following steps by using clean-text to the news articles before encoding them as an input of the models library \footnote{https://pypi.org/project/clean-text/}:

\begin{itemize}
    \item We lower tokens.
    \item We replace urls, emails, phones, numbers and currency symbols with specific tags.
    \item We fix the unicodes and remove all whitespaces.
\end{itemize}

\subsubsection{Training and Model Parameters}
We list the hyperparameters of each model and parameters for training them in Table~\ref{tab:hyperparams}. 
\begin{table}[t!]
    \centering
    \begin{tabular}{lll}
         \textbf{Model} & \textbf{Parameter Name} & \textbf{Value} \\
         3HAN & \textit{max length of words} & 20 \\
              & \textit{max length of headline} & 20 \\
              & \textit{max length of sentences} & 30 \\
              & \textit{pretrained model} & Glove \\
              & &Numberbatch \\
              & \textit{hidden dim} & 50 \\
              & \textit{word dropout} & 0.4 \\ 
              & \textit{sentence dropout} & 0.4 \\
              & \textit{title-body dropout} & 0.4 \\
              & \textit{max length of vocabulary} & 25000 \\
              & \textit{batch size} & 8 \\
              & \textit{epochs} & 100    \\        &&(with early stopping) \\
              & \textit{clipnorm} & 5 \\
              & \textit{learning rate} & 1e-3 \\

       HAN & \textit{max length of words} & 20 \\
              & \textit{max length of sentences} & 30 \\
              & \textit{pretrained model} & Glove \\
              & &Numberbatch \\
              & \textit{hidden dim} & 50 \\
              & \textit{word dropout} & 0.4 \\ 
              & \textit{sentence dropout} & 0.4 \\
              & \textit{max length of vocabulary} & 25000 \\
              & \textit{batch size} & 8 \\
              & \textit{epochs} & 100    \\        &&(with early stopping) \\
              & \textit{clipnorm} & 5 \\
              & \textit{learning rate} & 1e-3 \\
     SVM & \textit{max length of vocabulary} & 25000 \\
     RDEL & \textit{max length of vocabulary} & 25000 \\
           & \textit{epochs} & 100  \\
           &&(with early stopping) \\
            & \textit{clipnorm} & 5 \\
            & \textit{learning rate} & 1e-3 \\
     BERT   & batch size & 2 \\
      \&    & learning rate & 2e-5 \\
     Multitask    & epochs & 4 \\
       BERT   & max length of body & 512 \\
          & max lenght of title or statement & 128 \\
    \end{tabular}
    \caption{Training parameters and hyperparameters for each model.}
    \label{tab:hyperparams}
\end{table}

\end{document}